\documentclass[conference]{IEEEtran}
\IEEEoverridecommandlockouts
\usepackage{cite}
\usepackage{amsmath,amssymb,amsfonts}
\usepackage{algorithmic}
\usepackage{graphicx}
\usepackage{textcomp}
\usepackage{xcolor}
\def\BibTeX{{\rm B\ker=n-.05em{\sc i\kern-.025em b}\kern-.08em
    T\kern-.1667em\lower.7ex\hbox{E}\kern-.125emX}}
\begin{document}

\title{A multivariate water quality parameter prediction model using recurrent neural network\\}

\author{\IEEEauthorblockN {Dhruti Dheda}
\IEEEauthorblockA{\textit{School of Electrical and Information Engineering} \\
\textit{University of the Witwatersrand}\\
Johannesburg, South Africa \\
Dhruti.Dheda@students.wits.ac.za}
\and
\IEEEauthorblockN{Ling Cheng}
\IEEEauthorblockA{\textit{School of Electrical and Information Engineering} \\
\textit{University of the Witwatersrand}\\
Johannesburg, South Africa \\
Ling.Cheng@wits.ac.za}
}

\maketitle

\begin{abstract}
The global degradation of water resources is a matter of great concern, especially for the survival of humanity. The effective monitoring and management of existing water resources is necessary to achieve and maintain optimal water quality. The prediction of the quality of water resources will aid in the timely identification of possible problem areas and thus increase the efficiency of water management. The purpose of this research
is to develop a water quality prediction model based on water quality parameters through the application of a specialised recurrent neural network (RNN), Long Short-Term Memory (LSTM) and the use of historical water quality data over several years. Both multivariate single and multiple step LSTM models were developed, using a Rectified Linear Unit (ReLU) activation function and a Root Mean Square Propagation (RMSprop) optimiser was developed. The single step model attained an error of 0.01 mg/L, whilst the multiple step model achieved a Root Mean Squared Error (RMSE) of 0.227 mg/L.

\end{abstract}


\section{Introduction}
Pressure is placed on existing water resources through increased urbanisation, climate change and poor water infrastructure. Rivers are invaluable as inland water resources for agricultural needs, industrial and recreational purposes and for human consumption. Stress placed on existing water resources, especially
rivers necessitate the efficient management of these resources. To effectively manage water resources
the quality of the water needs to be continuously monitored [1]. However, the continuous accurate sampling and testing of water from existing water bodies can be very costly and tedious. For the most part, water quality is currently approximated through expensive and time-consuming laboratory analyses. The entire process from start to finish includes water sample collection from the relevant site, the correct storage and transportation of samples to the laboratory, chemical laboratory tests and analysis- which require a fair amount of time and the usage of expensive equipment, after which the water quality results will be found. During this long process there is a lot of room for error and inefficiency [2]. Water management can be made more efficient if data is analysed and water quality can be predicted beforehand [3]. To assist in this endeavour, this paper discusses the development of two predictive models for dissolved oxygen, a multivariate single step model and a multivariate multiple step model. The architecture and the accuracy of the models are compared. The current section, I section includes the introduction and the layout of the paper. Section II covers the research background in terms of river under study and the architecture of the neural networks under consideration. The III section covers the methodology followed to find the results that are presented and discussed in the IV section. The paper is concluded in section V where recommendations to the study are also covered.

\section{Background and preliminaries}

\subsection{Burnett River}

Burnett River is a river in south east Queensland, Australia. The Burnett river was named in 1847 after the surveyor J.C. Burnett, the first explorer of the river. It rises on the east of the Eastern Highlands, on the western slope of the Burnett Range. The course of the river is 435 km and it has a catchment area of 32,220 square km. The river takes the following course; it flows southwest to Eidsvold river and turns east at Mundubbera river and then northeast through Gayndah and Bundaberg rivers and eventually enters the Pacific Ocean at Burnett Heads. The Auburn and Boyne rivers as well as Barambah Creek are the main tributaries to the river [4]. 

\subsection{Artificial neural network}

Machine learning is a sub-field of artificial intelligence. Machine learning is a set of algorithms which analyse or deconstruct data and learns from the data; the algorithms then use what has been learnt (its experience) to find patterns of interest. Artificial Neural Networks (ANN) are a set of algorithms used in machine learning, which models the data with layers of neurons. ANN are based on the human brain. The brain’s ability to recognise, learn and memorise and still generalise patterns led to the field of study in algorithmic modelling of neural systems. The artificial neuron is modelled on the biological neuron. Each artificial neuron receives signals from other neurons or the environment; processes these signals and when excited transmits a signal to all the other connected neurons [5]. 

An ANN is composed of a layered network of artificial neurons. The architecture of the most general neural network consists one input layer, a middle section of one or more hidden layers and one output layer. The artificial neurons in each layer are interconnected to one another by numerical weights [6]. Each neuron collects incoming signals from which it calculates an overall net input signal respective to or as a function of the numerical weights. This net input signal then goes through the activation function. The activation function computes the output signal [5]. The excitement of an artificial neuron and the power of the output signal is controlled by the activation function [6]. 

\subsection{Recurrent neural network}

Deep Learning or Deep Neural Networks (DNN) is a subset of machine learning. DNN differs from conventional machine learning techniques in DNN’s ability to process data in its raw form. Previously, conventional machine learning techniques required the that a feature extractor of some sort be designed to alter the raw data into a feature vector, so that the learning system (like a classifier) could detect patterns in the input to the machine learning system. The DNN does not require the raw data to be altered prior to being inputted into the DNN system, as the DNN is capable of learning automatically from data that is unlabelled or unstructured [7].  The composition of the DNN allows for very complex functions to be learned. Neural networks learn and are optimised using the backpropagation algorithm. Backpropagation is a method of calculating gradients for each neuron or node in the network through recursive application of the chain rule [8]. The gradient value directs the way a neural network should alter the internal weights of the network thus allowing the network to learn. During back propagation, the gradient for each node is calculated in terms of the effects of the gradients of the nodes in the previous layer; implying that small weight adjustments in the previous layer (caused by small gradient values), will cause weight adjustments in the current layer to be even smaller and vice versa with bigger gradient values [7]. \

\begin{figure}[htbp]
\centerline{\includegraphics[scale=0.25]{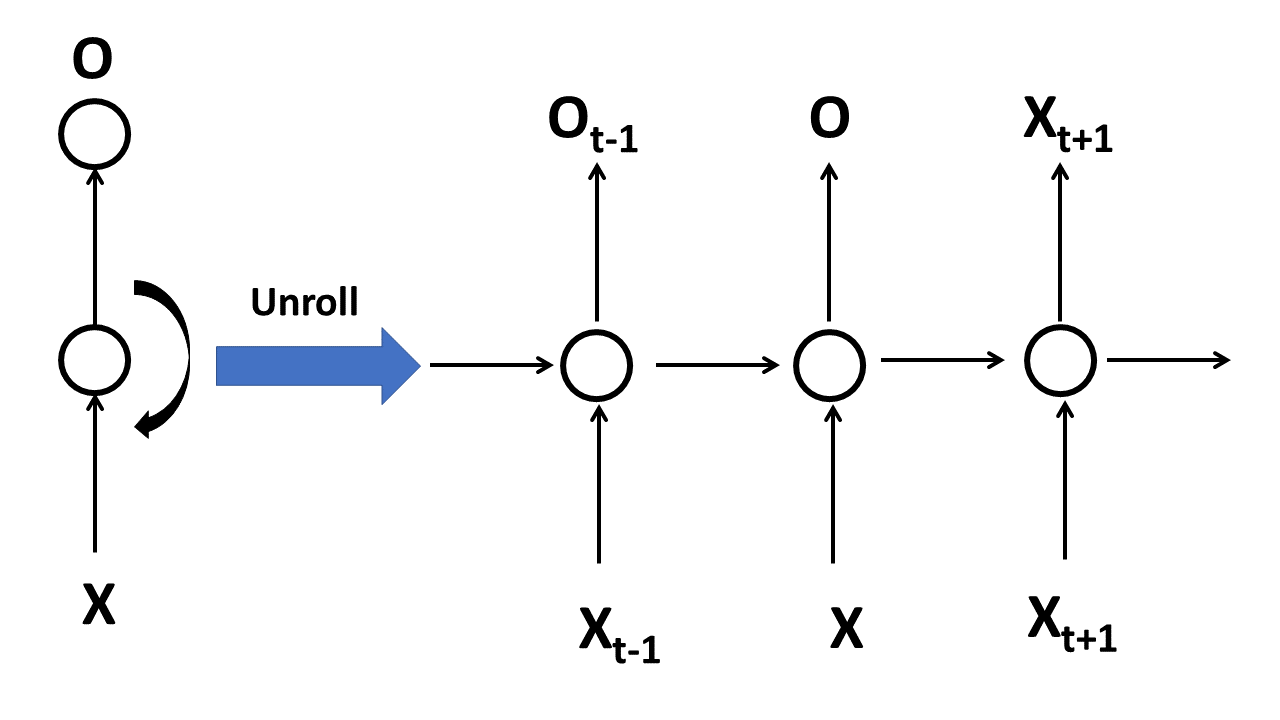}}
\caption{Single RNN (left) and same unrolled RNN where each layer has the same weights, biases and activation functions (right) adapted from[7]}
\label{fig}
\end{figure}

Most neural networks are feed forward neural networks that do not capture sequences nor model memory well. The Recurrent Neural Network (RNN) is a DNN that is specific to sequential input data. The simplest way to explain an RNN is to compare it to a feed forward neural network with loop that can pass information of previous states forward as shown in the diagram in the left in Figure 1. The looping mechanism of the RNN allows information to flow from one step to the next step, thus whilst processing the sequential data, the RNN retains information of the history of all the previous elements of the sequence that have been processed. This information which is a representation of the previous inputs is referred to as the hidden state [7]. Thus allowing RNNs to model memory. 

RNNs are trained using backpropagation through time- although this might seem counterintuitive at first. However, if the RNN is rolled out through time and considered at various discrete time steps, then each time step in the RNN can be thought of as a layer on its own as shown by the diagram of unrolled RNN on the right in Figure 1 [7].  When the architecture of the unrolled RNN, where the hidden state is shown by the rolled out individual hidden layers that each have the same weights, biases and activation functions, is considered it becomes plausible to apply backpropagation through time to the forward computation of the network to calculate the gradients of the nodes [7]. 

However, despite the dynamism of RNNs, training RNNs with backpropagated gradients has its negatives. When training RNNs using backpropagation, the gradients either shrink or grow at each time step- due to the gradients at each layer being calculated with respect to the gradients of the previous layer as discussed above. Hence over multiple time steps, the gradients will either vanish or explode [9]. This results in the earlier layers of the network failing to do any learning as internal weights are barely being adjusted due to extremely small gradient values- vanishing gradient. Due to the vanishing gradients the RNN does not learn the long-range dependencies across time steps; resulting in RNNs being able to learn short term but not long term dependencies [8]. 

Numerous attempts have been made to address the difficulties faced when training RNNs. The case of vanishing gradients was first successfully addressed by Hochreiter \& Schmidhuber in 1997 [10]. Hochreiter \& Schmidhuber developed the Long Short-Term Memory (LSTM) architecture which became the standard way of dealing with vanishing gradients [7].  The issue of exploding gradients is relatively simpler to deal with when compared to that of the vanishing gradient problem. A technique known as gradient clipping, which entails shrinking the gradient when norms exceed a threshold, is easy to implement [7]. 

\subsection{Long short-term memory}

LSTMs are capable of learning long term dependencies using mechanisms called gates. Each LSTM cell has a cell state, considered to be the memory of the network that transfers relative information all the way down the sequence chain. Information is added to or disregarded from the cell state via the use of gates. The gates can learn which information is important to keep or disregard, in doing so relevant information is used to make predictions. There are three different gates, the forget, input and output gates. Each gate contains sigmoid activations [11].

The forget gate determines which information should be retained and excluded. The previous hidden state and the current input to the LSTM cell is multiplied by the weight of the forget gate and added to one another and then passed through the sigmoid activation [11] as shown in (1). It should be noted that each gate has a different set of weights. The equations expressed below were adapted from [12]

\begin{equation}
f_t= \sigma(W_f {S_{(t-1)}} + W_f X_t)   \label{eq}
\end{equation}

where $f_t$ is the forget gate, $W_f$ is the weight of the forget gate, $S_t-1$ is the previous hidden state, $X_t$ is the input state and σ is the sigmoid activation. 
The input gate is used to update the cell state when the previous hidden state and current input is passed through sigmoid activation that decides which values will be updated in accordance [11] to (2).

\begin{equation}
I_t= \sigma(W_i {S_{(t-1)}} + W_i X_t)   \label{eqA}
\end{equation}
where $I_t$ is the input gate, $W_i$ is the weight of the input gate. 
Also, an intermediate cell state is calculated, where the previous hidden state and the current input is passed through a tanh function to regulate the network [11] as expressed in (3). 

\begin{equation}
c_t= tanh(W_c {S_{(t-1)}} + W_c X_t)   \label{eqB}
\end{equation}
where $c_t$ is the intermediate cell state, $W_c$ is the weight of the intermediate cell state. 

The output of the intermediate cell state is then multiplied by the output of the input gate, in this way the sigmoid output determines what should be retained from the tanh output; this is then added (through pointwise addition) to the product of the output of the forget gate and the previous cell state to calculate the current cell state [11] as expressed in (4). 

\begin{equation}
C_t=(I_t c_t)+(f_t {C_{(t-1)}} )      \label{eqC}
\end{equation}
where $C_t$ is the current cell state, $C_t-1$ is the previous state.

The output gate then determines what the next hidden state should be using the current cell state. First the previous hidden state and the current input is passed through sigmoid activation [11] as shown in (5).

\begin{equation}
O_t=\sigma(W_o {S_{(t-1)}} + W_o X_t)   \label{eqD}
\end{equation}
where $O_t$ is the output gate, $W_o$ is the weight of the output gate.

The current cell state is then passed through tanh activation and multiplied by the output of the output gate to calculate the hidden state as shown in (6) as this determines the information that the hidden state will carry to the next time step [11].

\begin{equation}
h_t=O_t×tanh(C_t)   \label{eqE}
\end{equation}
where $h_t$ is the hidden state.

As previously stated, RNNs are good for processing sequence data for predictions but can only learn short term but not long-term dependencies across time steps. Through the use of gates, LSTMs are capable of learning long-term dependencies across time steps and thus short-term memory is not an issue for LSTMs. LSTMs are better to use when modelling longer sequences with long term dependencies [11]. Although RNNs suffer from short term memory, they have the benefit of training faster and using less computational as there are fewer tensor operations to complete. Thus, the decision of which one is better would depend on the data and the purpose for which the data is being modelled. LSTMs have proven to be more effective than traditional RNNs in recent years [8]. 

The architecture of the LSTM can be more clearly illustrated in Figure 2 [12], which shows the internal structure of an LSTM recurrent network cell, in accordance to the equations previously expressed.  

\begin{figure}[htbp]
\centerline{\includegraphics[scale=0.25]{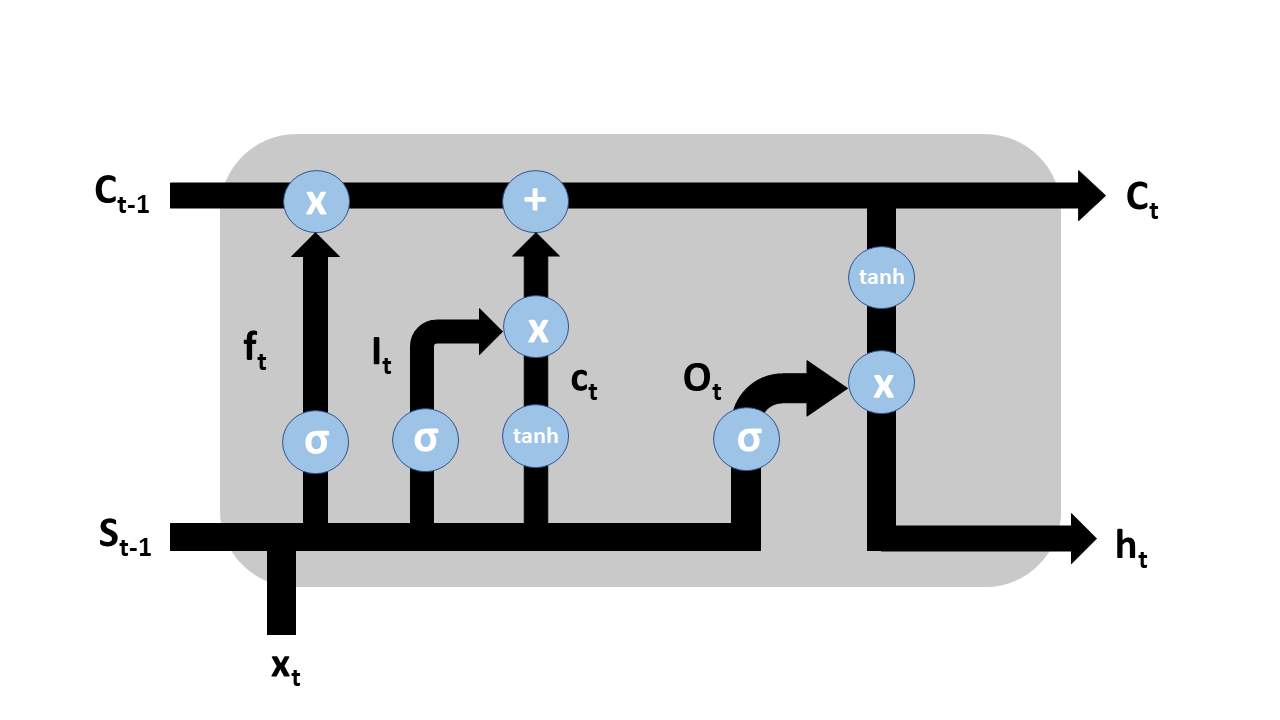}}
\caption{Internal structure of a LSTM recurrent network cell adapted from [12]}
\label{figa}
\end{figure}

\section{Multivariate Water Quality Parameter Prediction Scheme}
Historical water quality for the Burnett River was provided through a publicly available data source from the Ambient Estuarine Water Quality Monitoring Programme from the Queensland Government open data portal for the years from the beginning of 2013 to the end of 2019, with observations of water quality parameters given every 30 minutes of each hour for every day in the year [13].  This comes to 48 observation per day and roughly 17250 recorded observations for each water quality parameter per year.  

The following processes that formed the methodology that was applied is described below and is illustrated in Figure 3 for ease of understanding. 
\begin{figure}[htbp]
\centerline{\includegraphics[scale=0.40]{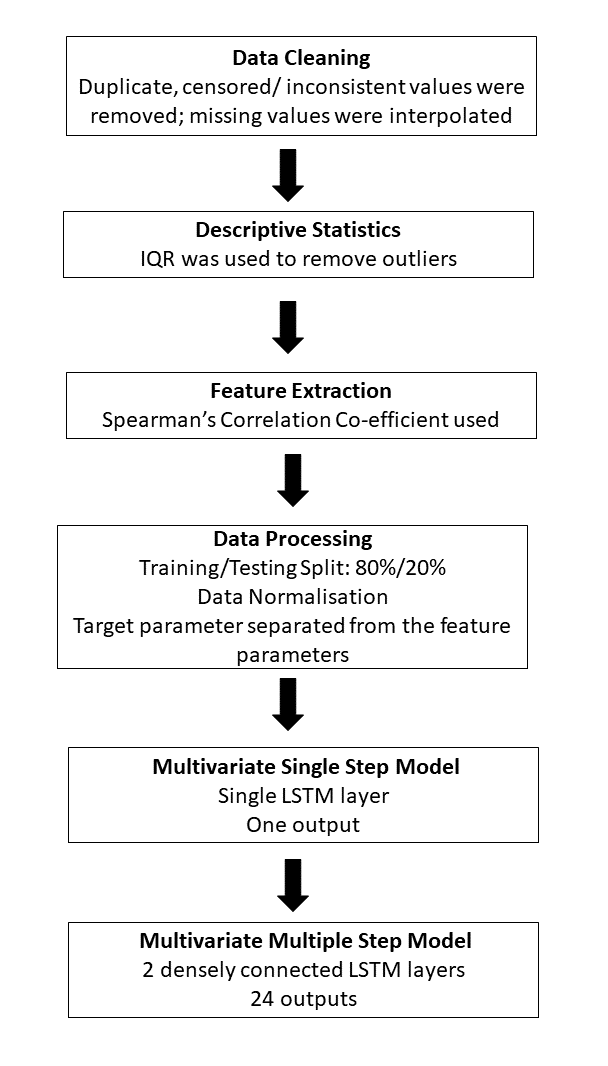}}
\caption{Schematic diagram of applied methodology}
\label{figb}
\end{figure}

The data included water quality parameters such as water temperature, pH value, electrical conductivity, dissolved oxygen and turbidity. The data was cleaned to remove any censored or inconsistent observations; duplicate values were also removed. The sequential data was then checked for any observations that were not in line with the regular time intervals, observations at inconsistent time intervals were removed and missing values were found through interpolation. Statistical analysis over the data was then carried out. The minimum, maximum, mean, standard deviation, median and first and third quartile values of each water quality parameter were calculated to gain greater insight of the data. It was easier to remove outliers after the upper and lower boundaries of each parameter was found using the interquartile range (IQR). Outliers were then removed. Data cleaning was completed. 

The strength of the relationships between the water quality parameters was evaluated using the correlation coefficients. Correlation is the measure of association between two variables, often between the values of -1 and 1. Variables can have a positive association, meaning that as one variable increases so does the other variable. They may have a negative association in which one variable increases whilst the other variable decreases. A neutral correlation, close to the value of zero implies that there is no relationship between the variables. It is important to gauge the correlation between the variables, to determine which variables may be relevant as inputs for the development of a predictive model. Hence a strong correlation between the two variables implies that they can be used to build a model. A commonly used correlation method is Spearman’s Correlation, named after the founder, Charles Spearman (as are most correlation methods) [14]. After the correlation coefficients were found, variables which could be used as inputs for the model were chosen. 

Two models were built. The first model was a multivariate single step model that predicted only one future value of the target water quality parameter using several water quality parameters (features) for the prediction. The second model was multivariate multiple step model which used several water quality parameters to predict many values of the target water quality parameter in the future. 

The architecture of the single step model employed was a sequential model was a basic LSTM model with a single LSTM layer with 3 input nodes into the input layer and one output layer that returned a single value predicted 12 hours into the future. The the optimiser used was RMSprop. 

The architecture of the multiple step model employed was a sequential model was a LSTM model with two densely connected LSTM layers with 32 input nodes into an input layer and one output layer that returned 24 values of the target water quality parameter predicted for every observation made every 30 minutes for the next 12 hours into the future. The activation function used was ReLU activation function and the optimiser used was RMSprop.  

The activation function that was used was the Rectified Linear Unit (ReLU) activation function. The ReLU is a piecewise linear function or a hinge function as it can be linear for some of the input but nonlinear for the rest depending on whether the input is positive or negative. For values greater than zero, the ReLU function is linear, implying that the properties of a linear function are retained for training neural networks with backpropagation. For values less than zero, for negative values, ReLU behaves like a nonlinear function and outputs a zero [11]. In recent times, ReLU has become the trusted default activation function when developing most types ANN. The advantages of the ReLU function were highlighted by Xavier Glorot et.al [15] in the paper, ‘Deep Sparse Rectifier Neural Networks’; these include computational simplicity, representational sparsity and linear behaviour. Unlike the other popular activation functions, the sigmoid and tanh function, the ReLU does not require the computation of an exponential function in activations and is hence easier and cheaper to implement [15]. Representational sparsity occurs when negative inputs can output true zero values, which differs to the sigmoid and tanh function which approximates a value close to zero but not a true zero value. By outputting a true zero value from negative inputs, the ReLU activation function allows hidden layers in the neural network to be activated and hence to contain at least one true zero value [11]. Neural networks that are linear or close to linear are easier to optimise than other types of neural networks. In general, the ReLU activation function appears and behaves like a linear activation function [11]. Due to the linearity of this activation function, the gradients remain proportional to the node activations and thus neural networks trained with ReLU activation function do not have to deal with the issue of vanishing gradients [15].

The chosen optimiser was the Root Mean Square Propagation (RMSprop) optimiser. RMSprop is notorious for being an unpublished optimiser which was suggested by Geoff Hinton in a lecture. RMSprop is an adaptive learning rate method; a good default value for the learning rate is 0.001. This optimiser was developed to address the problem of drastically diminishing learning rates that were observed using the Adagrad optimiser [16]. 

The cleaned data was split into training data and testing data, 80- 20 percent respectively. The target value or output variable was then separated from the features or input values in the training and test sets. As the features had different ranges and scales, the features were normalised. It is possible for a model to converge without the normalisation of data but training such a model is much more difficult and the model is dependent on the input unit choice. The mean and standard deviation were the statistics used to normalise the data in both the training and test dataset. The normalised training set data was used to train the model.

The model was then trained with the normalised data for thousand epochs. An epoch is one learning iteration. A learning iteration consists of two phases, a feedforward pass and back propagation. For each training pattern, the feedforward pass calculates the output value of the neural network. When an error signal is propagated back from the output layer to the input layer, it is referred to as back propagation. The weights in the neural network are then adjusted as function of the back propagated error signal [5].

The training and validation accuracy were measured using the Mean Absolute Error (MAE). The model was trained for a thousand epochs. Before a thousand epochs was reached, the graph shows very little improvement in validation accuracy in terms of the MAE error and hence a thousand epochs were not necessary. Early stopping of the epoch was implemented. Early stopping tested a training condition after each epoch. If a certain number of epochs passed without showing any improvement in validation accuracy, the training of the model was automatically stopped. This is how the most appropriate number of epochs was found. 

The model was then trained again using the appropriate number of epochs and the MAE on the validation set was recorded. The model was then employed on the normalised test set and the MAE of the test set was found. Finally, the model was used to predict target values using the data in the normalised test set. 

\section{Results and Discussion}
Both the multivariate single and multiple step model was used to predict the value of Dissolved Oxygen, twelve hours into the future. 
The amount of free non-bonded, non-compound oxygen present in freshwater is referred to as dissolved oxygen. Dissolved oxygen is a significant parameter when evaluating the water quality of a freshwater body, owing to the effect that dissolved oxygen levels have on the river dwelling organisms. Freshwater dwelling organisms, such as fish, invertebrates, plants and bacteria use dissolved oxygen in water in a manner similar to how organisms on land use oxygen. The amount of oxygen required by each of these organisms will differ from organism to organism. Fish can receive oxygen through their gills for respiration, whilst plant life might require dissolved oxygen for respiration light required for photosynthesis is not available. Levels of dissolved oxygen are constantly affected by aeration and diffusion and as such dissolved oxygen levels fluctuate easily whilst the river water equilibrates towards a hundred percent air saturation (equilibrium point for gases in water as the gas molecules diffuse between the atmosphere and the surface of the water). If the dissolved oxygen level is either too high or too low, it can adversely affect water quality and harm aquatic life; hence monitoring and possibly predicting dissolved oxygen levels are of paramount importance [17].  

There were several features (water quality parameters) provided by the data but only some of them correlated to dissolved oxygen. Table 1 shows the Spearman’s Correlation Coefficient calculated for each water quality parameter with regards to dissolved oxygen.
\begin{table}[htbp]
\caption{Spearman’s Correlation Coefficient with regards to dissolved oxygen}
\begin{center}
\begin{tabular}{c|c}
\hline
\textbf{Water Quality Parameters}&\multicolumn{1}{|c}{\textbf{Spearman's Correlation Coefficient }} \\
\hline
pH& 0.436\\
Temperature& -0.436  \\
Electrical Conductivity& -0.218   \\
Dissolved Oxygen& 1.000   \\
Turbidity& -0.114   \\
\hline
\end{tabular}
\label{tab1}
\end{center}
\end{table}

As shown in Table 1, dissolved oxygen has a strong positive relationship with pH and strong negative relationship with temperature. Hence as the temperature of the water increases, the dissolved oxygen level of the water decreases. This observation is consistent with literature which states that the solubility of oxygen will decrease as the temperature of the water increases as warmer water will require a lower amount of dissolved oxygen to reach a hundred percent air saturation than colder water [17]. As the dissolved oxygen level in the water increases the pH level increases.

The relationships between dissolved oxygen and the other water quality parameters mentioned in Table 1 are either weak or almost too trivial to be considered, such as relationships that dissolved oxygen shares with electrical conductivity and turbidity, quantified as very small correlation coefficients. Water quality parameters that share both weak and trivial relationships with pH were disregarded when building the model. Temperature and pH were the features used to the build the model. The data for the two features and the target output, pH was given for the period from 2013 to 2019. 

Table 2 shows the overall statistics of the water quality parameters including the mean, standard deviation, the minimum and maximum values as well as the spread of the data across the 25, 50 and 75 percent boundaries. 
\begin{table}[htbp]
\caption{Descriptive statics of the chosen water quality parameters}
\begin{center}
\begin{tabular}{c|c|c|c|c|c|c|c}
\hline
\textbf{Parameters}&{\textbf{Mean}}&{\textbf{Std Dev}}&{\textbf{Min}}&{\textbf{25}}&{\textbf{50}}&{\textbf{75}}&{\textbf{Max}} \\
\cline{2-4} 
\hline
Temp&24&3.7&15.7&20.9&24.9&27.9&31.4\\
DO&6.6&0.6&4.8&6.2&6.6&7.0&9.4  \\
pH&7.8&0.1&7.3&7.7&7.8&7.9&8.3 \\
\hline
\end{tabular}
\label{tab2}
\end{center}
\end{table}
As can be seen from Table 2, the ranges and scales of the features differ from one another and hence it was necessary to normalise the data using the mean and standard deviation. The normalised data was used to train the model. 
The pH value ranges from 7.3 to 8.3 with a mean pH value of 7.8. In general, depending on the surrounding soil, the pH value of freshwater ponds, lakes and streams range from 6–-8 [18]. The pH values in this dataset fall within this range. The water (Temp) temperature (\textdegree C) ranges from 15.7\textdegree C to 31.4\textdegree C with a mean water temperature of 24.5\textdegree C. Typical water temperatures will be dependent on the type of water body and the surrounding environment. Streams and rivers experience greater temperature fluctuations than oceans and lakes [19]. The dissolved oxygen (DO) level ranges from 4.3 mg/L to 9.4mg/L with a mean dissolved oxygen level of 6.6 mg/L. As previously stated, dissolved oxygen levels fluctuate easily; it is possible for a river to show daily fluctuations of 3 mg/L of dissolved fluctuations. Depending on the physical location of the river and its surrounding conditions the level of dissolved oxygen can range from 6–-9 mg/L. The couple of the values in this dataset may deviate from this range, possibly owing to human error, however most of the values fall into range as shown from the mean (6.6 mg/L) and the lower (6.2 mg/L) and upper (7.0 mg/L) quartiles [17]. 

\begin{figure}[htbp]
\centerline{\includegraphics[scale=0.35]{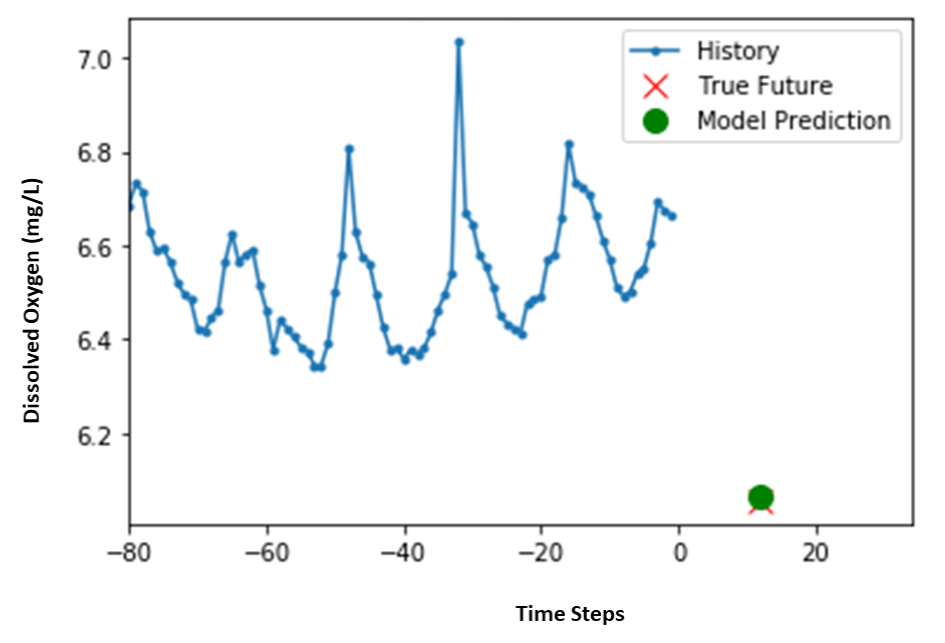}}
\caption{Multivariate single step prediction model for dissolved oxygen 12 hours ahead}
\label{figc}
\end{figure}

The multivariate single step model uses the Root Mean Square Propagation optimiser, with a single LSTM layer. The optimal number of epochs for the prediction of dissolved oxygen was found when certain number of epochs passed without showing any improvement in validation accuracy, the training of the model was terminated and the number of epochs that had passed was the optimal number of epochs for the training of the model. The optimal number of epochs for the training of model was found to be 400 epochs. The model was trained again, using 400 epochs. To determine how well the model generalises, it was tested using the testing set which was not used to train the model. Figure 4 shows that the trained model was able to predict the dissolved oxygen quite well, 12 hours into the future. As can be seen in Figure 4, the green point for the model prediction at 6.08 mg/L meets the red cross for true future at 6.07 mg/L. The error between the points are 0.01 mg/L. 

The multivariate multiple step model uses the rectified linear activation function, the Root Mean Square Propagation optimiser, with two densely connected LSTM layers. The optimal number of epochs for the training of model was found to be 1000 epochs. The model was tested using the testing set which was not used to train the model. Figure 5 shows that the trained model was able to predict 24 values of dissolved oxygen, 12 hours in the future, one value for every 30 minutes. As can be seen in Figure 5 , the red points are the dissolved oxygen predictions made by the model and the blue points are the true future. The model predicts reasonably well with a root mean squared error of 0.227 mg/L.
\begin{figure}[htbp]
\centerline{\includegraphics[scale=0.29]{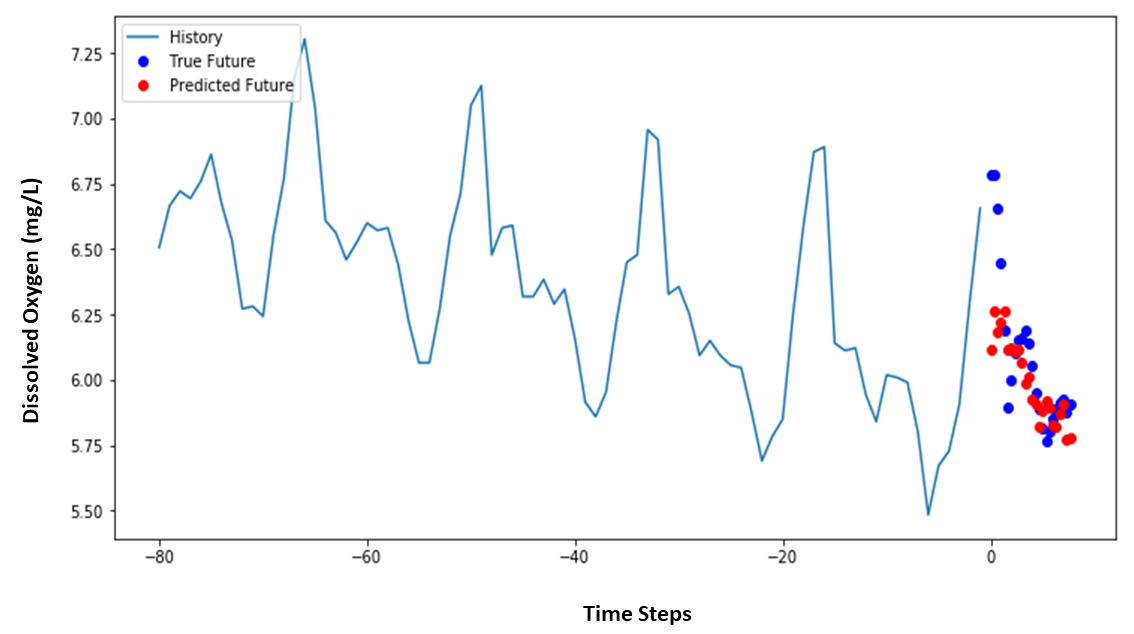}}
\caption{Multivariate multiple step prediction model for dissolved oxygen 12 hours ahead}
\label{figd}
\end{figure}

Both models are multivariate as both the models made use of pH and water temperature to predict the level of dissolved oxygen. The architecture of the models differed greatly. The single step model only had one LSTM layer with 3 input nodes, whilst the multiple step model had two densely connected LSTM layers with 32 input nodes into the first layer and 16 nodes into the second layer. Both models had to predict the dissolved oxygen level twelve hours into the future. The single step model optimally trained from fewer epochs than the multiple step model, this may be owing to the smaller architecture of the single step model. The multiple step model took longer to train and would train poorly with one a single LSTM layer. The single step model produced a small error of 0.01 mg/L, when compared to the multiple step model which had an RMSE of 0.227 mg/L. From these results it can be concluded the single step model predicts the dissolved oxygen value with greater accuracy than the multiple step model. However, it must also be noted that as the multiple step model has to predict the 24 dissolved oxygen values, 12 hours ahead as opposed to the single value predicted by the single step model, the possibility of errors occurring increases. 

\section{Conclusion}

This paper has shown the development of a water quality prediction model based on other water quality parameters through the application of specialised recurrent neural network (RNN), Long Short Term Memory (LSTM) and the use of water quality data. Both a multivariate single and multiple step LSTM models were developed, one with a single LSTM layer and the other with two densely connected LSTM layers, using a Rectified Linear Unit (ReLU) activation function and a Root Mean Square Propagation (RMSprop) optimiser. The optimal number of epochs for the training of the single step dissolved oxygen predictive model was found to be 400 epochs. The model predicted quite well with an error of 0.01 mg/L. The optimal number of epochs for the training of the multiple step dissolved oxygen predictive model was found to be 1000 epochs. The model achieved a root mean squared error of 0.227 mg/L. The model predicted reasonably well. The prediction of the quality of water resources, in terms of water quality parameters through the application of deep neural networks, a subfield of machine learning will aid in the timely identification of potential problem areas and thus increase the efficiency of water management- a necessity in a global water stressed future.  This paper would recommend that predictive LSTM models be developed for each of the other water quality parameters given; each model only utilising the water quality parameters that share the greatest correlation with the target that the model aims to predict. A comparative study of the predictive models should be conducted for similarities and differences.

\vspace{12pt}

\end{document}